\documentclass[11pt, oneside]{article}  

\usepackage{geometry}  
\usepackage{cmbright}
\usepackage{cite}
\usepackage{graphicx}				
\usepackage{tcolorbox}
\usepackage{amssymb}
\usepackage{longtable}
\usepackage{amsmath}
\usepackage{booktabs}
\usepackage{url}
\usepackage{color}
\definecolor{c1}{rgb}{0.12, 0.56, 1.0}
\usepackage{xspace}
\usepackage{xcolor}
\usepackage{caption}
\usepackage{subcaption}
\usepackage{authblk} 

\usepackage{sectsty} 
\definecolor{cool_blue}{RGB}{24, 132, 193}
\sectionfont{\color{c1}\selectfont}
\subsectionfont{\color{c1}\selectfont}
\subsubsectionfont{\color{c1}\selectfont}
\subparagraphfont{\color{gray}\selectfont}

\definecolor{fruitpushorange}{RGB}{255, 127, 0}

\usepackage{tcolorbox} 
\tcbuselibrary{breakable}
\newtcolorbox[auto counter
]{mybox}[2][]{
title=Box~\thetcbcounter: #2,#1,
colback=white,
colframe=cool_blue,
fonttitle=\bfseries,
parbox=false
}

\usepackage{soul}

\title{
Bi-objective trail-planning for a robot team orienteering in a hazardous environment
}
\author[1]{Cory M. Simon$^*$}
\author[2]{Jeffrey Richley}
\author[2]{Lucas Overbey$^\ddagger$}
\author[2]{Darleen Perez-Lavin$^\dagger$}
\affil[1]{School of Chemical, Biological, and Environmental Engineering. Oregon State University. Corvallis, OR. USA.}
\affil[$^*$]{\texttt{cory.simon@oregonstate.edu}}
\affil[2]{Naval Information Warfare Center Atlantic. Charleston, SC. USA.}
\affil[$^\ddagger$]{\texttt{lucas.a.overbey.civ@us.navy.mil}}
\affil[$^\dagger$]{\texttt{darleen.s.perez-lavin.civ@us.navy.mil}}

\begin{document}
\maketitle

\begin{abstract}
Teams of mobile [aerial, ground, or aquatic] robots have applications in resource delivery, patrolling, information-gathering, agriculture, forest fire fighting, chemical plume source localization and mapping, and search-and-rescue.
Robot teams traversing hazardous environments---with e.g.\ rough terrain or seas, strong winds, or adversaries capable of attacking or capturing robots---should plan and coordinate their trails in consideration of risks of disablement, destruction, or capture.
Specifically, the robots should take the safest trails, coordinate their trails to cooperatively achieve the team-level objective with robustness to robot failures, and balance the reward from visiting locations against risks of robot losses.

Herein, we consider bi-objective trail-planning for a mobile team of robots orienteering in a hazardous environment.
The hazardous environment is abstracted as a directed graph whose arcs, when traversed by a robot, present known probabilities of survival.
Each node of the graph offers a reward to the team if visited by a robot (which e.g.\ delivers a good to or images the node).
We wish to search for the Pareto-optimal robot-team trail plans that maximize two [conflicting] team objectives: the expected (i) team reward and (ii) number of robots that survive the mission. 
A human decision-maker can then select trail plans that balance, according to their values, reward and robot survival.
We implement ant colony optimization, guided by heuristics, to search for the Pareto-optimal set of robot team trail plans. 
As a case study, we illustrate with an information-gathering mission in an art museum.

\end{abstract}

\clearpage

\section{Introduction}
\subsection{Applications of a team of mobile robots}
Mobile [aerial \cite{leutenegger2016flying}, ground \cite{chung2016wheeled}, or aquatic \cite{choi2016underwater}] robots equipped with sensors, actuators, and/or cargo have applications in agriculture (e.g.\ planting and harvesting crops, spraying pesticide, monitoring crop health, destroying weeds) \cite{santos2020path,bawden2017robot,mcallister2018multi}, commerce (e.g.\ order fulfillment in warehouses) \cite{wurman2008coordinating}, the delivery of goods \cite{coelho2014thirty}, search-and-rescue \cite{queralta2020collaborative,rouvcek2020darpa}, chemical, biological, radiological, or nuclear incident response (e.g.\ safely localizing the source(s) and mapping the distribution of the hazard) \cite{murphy2012projected,hutchinson2019unmanned}, environmental monitoring \cite{dunbabin2012robots,hernandez2012mobile,yuan2020maritime}, safety monitoring in industrial chemical plants \cite{soldan2014towards,francis2022gas}, forest fire monitoring and fighting \cite{merino2012unmanned}, target tracking \cite{robin2016multi}, and military surveillance and reconnaissance. 

Often, we wish for a team of mobile robots to coordinate their paths in an environment to cooperatively achieve a shared, team-level objective \cite{parker1995design,parker2007distributed}.
Compared to a single robot, a team can increase spatial coverage, decrease the time to achieve the objectives, and make achievement of the objectives robust to the failure of robots \cite{schranz2020swarm,brambilla2013swarm}.

For example, consider the (NP-hard) team orienteering \cite{golden1987orienteering} problem (TOP) \cite{chao1996team,gunawan2016orienteering,vansteenwegen2011orienteering}. 
The environment, in which a team of robots are mobile, is modeled as a graph (nodes: locations; edges: spatial connections between locations). Each node offers a reward to the team if visited by a robot.
The TOP is to plan the paths of the robots between a source and destination node, subject to a per-robot travel budget, to accumulate the most rewards from the graph as a team. The TOP can be formulated as an integer program. Loosely, the orienteering problem combines the classic knapsack problem (selecting the nodes from which to collect rewards, under the travel budget) and traveling salesman problem (finding the shortest path that visits these nodes) \cite{vansteenwegen2011orienteering}.

\subsection{Teams of mobile robots orienteering in risky environments} 
In some applications, robots move in a hazardous environment \cite{trevelyan2016robotics} and incur risks of failure, destruction, disablement, and/or capture. 
The hazards could originate from dangerous terrain, rough seas, strong winds, heat, radiation, corrosive chemicals, or mines---or an adversary with the capability to attack, disable, destroy, or capture robots \cite{agmon2017robotic}. 

Robots traversing a hazardous environment should plan and coordinate their trails in consideration of risks of failure.
First, robots should take the safest trails to visit their destination(s). 
Second, the robots should coordinate their trails to make achievement of the team objective resilient to robot failures \cite{zhou2021multi}. 
A \emph{resilient} team of robots \cite{prorok2021beyond}
(i) anticipates failures and makes risk-aware plans that endow the team with \emph{robustness}---the ability to withstand failures with minimal concession of the objective,
and/or
(ii) adapts their trail plans during the mission, in response to realized robot failures, to recoup the anticipated loss in the objective. 
Third, the trail plans must balance the rewards gained from visiting different locations against the risks incurred by the robots to reach those locations.

Models and algorithms have been developed for path-planning for robot teams orienteering in risky environments abstracted as graphs \cite{zhou2021multi}. 
In the Team Surviving Orienteers problem (TSOP) \cite{jorgensen2018team,jorgensen2017matroid,jorgensen2024matroid}, each node of the graph offers a reward to the team when visited by a robot, and each edge-traversal by a robot incurs a probability of destruction. 
The objective in the TSOP is to plan the paths of the robots (from a source to destination node) to maximize the expected team reward under the constraint that each robot survives the mission with a probability above a set threshold. 
(In the \emph{offline} setting, the paths of the robots are set at the beginning of the mission, then followed without adaptation. In the \emph{online} setting, the paths are updated during the mission in response to realized robot failures.)
Relatedly, the Foraging Route with the Maximum Expected Utility problem \cite{di2022foraging} is to plan the foraging route of a robot collecting rewards in a hazardous environment, but the rewards are lost if the robot is destroyed before returning to the source node to deposit the goods it collected.
In the Robust Multiple-path Orienteering Problem (RMOP) \cite{shi2023robust}, similarly, each node gives a reward to the team only if a robot visits it \emph{and} survives the mission; the paths of the $K$ robots are planned to maximize the team reward under the worst-case attacks of $\alpha<K$ of the robots by an adversary. 
The offline version of RMOP constitutes a two-stage, sequential, one-step game with perfect information: (1) the robot team chooses a set of paths then (2) the adversary, knowing these paths, chooses the subset of robots to attack and destroy. 
The optimal path plans for the robots must trade (i) redundancy in the nodes visited, to give robustness against attacks, and (ii) coverage of many nodes, to collect many rewards.
Other work involving robot path-planning in hazardous environments includes 
maximizing coverage of an area containing threats to robots \cite{korngut2023multi,yehoshua2016robotic}, 
handling adversarial attacks on the sensors of the robots \cite{liu2021distributed,zhou2022distributed,mayya2022adaptive,zhou2018resilient}, 
gathering information in an environment with unknown hazards \cite{schwager2017multi},
finding the optimal formation for a robot team \cite{shapira2015path},
and 
multi-robot patrolling under adversarial attacks \cite{huang2019survey}.

\subsection{Our contribution}
\begin{figure}[h!]
    \centering
     \begin{subfigure}[b]{0.62\textwidth}
    	\includegraphics[width=\textwidth]{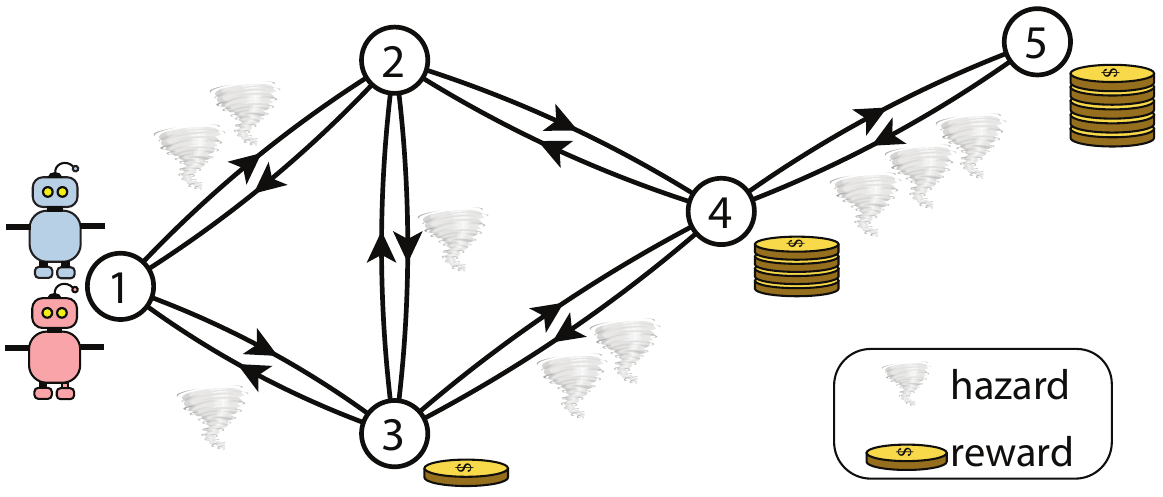}
	\caption{} \label{fig:overview}
    \end{subfigure}
    \begin{subfigure}[b]{0.66\textwidth}
    	\includegraphics[width=\textwidth]{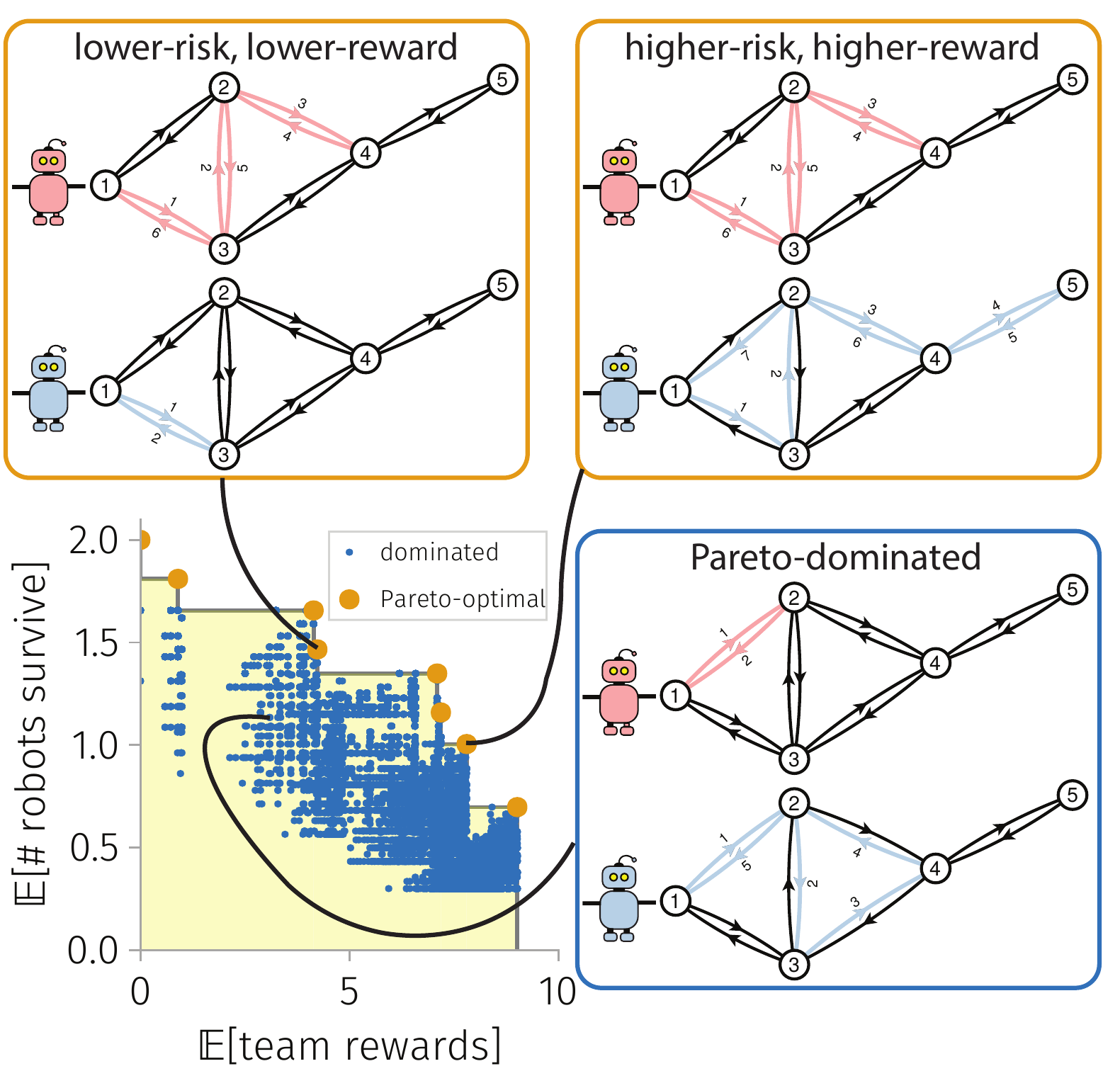}
	\caption{} \label{fig:pareto_optimal}
    \end{subfigure}
    \caption{
      The bi-objective team orienteering in hazardous environments (BOTHE) problem.
      (a) A team of robots are mobile on a directed graph whose 
      nodes offer a reward to the team if visited by a robot and 
      arcs present a probability of destruction to robots that traverse them (tornado: 1/10 probability of destruction). The task is to plan the trails of the robots to maximize the expected reward collected and the expected number of robots that survive.
      (b) Pareto-optimal and -dominated robot-team trail plans scattered in objective space, with two Pareto-optimal plans and one Pareto-dominated plan shown.}
\end{figure}

Herein, we frame the bi-objective team orienteering in hazardous environments (BOTOHE) problem---a variant of the offline TSOP \cite{jorgensen2018team,jorgensen2017matroid,jorgensen2024matroid}.
A team of robots are mobile in a hazardous environment, modeled as a directed graph whose arcs present known probabilities of destruction to robots traversing them.
Each node of the graph offers a reward to the team if visited by a robot.
The BOTHE problem is to plan the closed trails of the robots to maximize two team-level objectives: the expected
(1) rewards accumulated by the team via visiting nodes and
(2) number of robots that survive the mission. 
(We focus on the offline setting, owing to the lack of communication with/between robots after the mission executes.)
See Fig.~\ref{fig:overview}.

Three interesting features of the BOTOHE are: 
(1) for the survival objective, robots risk-aversely a) avoid visiting dangerous subgraphs despite rewards offered and b) take the safest closed trails to visit the nodes assigned to them;
(2) for the reward objective, the robots daringly a) visit dangerous subgraphs to attempt collection of the rewards offered, b) visit the lower-risk and higher-reward subgraphs earlier in the mission, and c) build node-visit redundancy into their trail plans to make the team-reward robust to the loss of robots during the mission; and
(3) comparing (1a) and (2a), the two objectives are inherently conflicting, as the robots must risk their survival while taking their trails to visit nodes and collect rewards.

To handle the conflict between the two objectives in the BOTOHE, we search for the Pareto-optimal set \cite{pardalos2017non,branke2008multiobjective} of robot-team trail plans. By definition, a Pareto-optimal trail plan cannot be altered to give higher expected rewards without lowering the expected number of robots that survive---and vice versa. 
Fig.~\ref{fig:pareto_optimal} illustrates.
We then can present the Pareto-optimal set to a human decision-maker to select the team trail plan that strikes some tradeoff between the two objectives, based on their value of rewards vs. robot survival. 

We employ a bi-objective ant colony optimization algorithm \cite{iredi2001bi}, guided by heuristics, to search for the Pareto-optimal set of robot-team trail plans.
For illustration, we solve and analyze a BOTOHE problem instance for information-gathering in an art museum.
Through ablation studies, we quantify the contribution of the greedy heuristics and pheromone to the search efficiency. 

\section{The bi-objective team orienteering in hazardous environments (BOTHE) problem}


\subsection{Problem setup}
Here, we frame the Bi-Objective robot-Team Orienteering in Hazardous Environments (BOTOHE) problem.

A homogenous team of $K$ robots are mobile in an environment modeled as a directed graph $G=(\mathcal{V}, \mathcal{E})$. Each node $v \in \mathcal{V}$ represents a location in the environment (e.g., a room in a building). Each arc $(v, v^\prime) \in\mathcal{E}$, an ordered pair of distinct nodes, represents a one-way, direct spatial connection (e.g.\ a door- or hall-way) to travel from node $v$ to node $v^\prime$. 
\emph{Mobility} of the robots implies they may walk on the graph $G$, i.e. sequentially hop from a node $v$ to another node $v^\prime$ via traversing arc $(v, v^\prime)\in\mathcal{E}$.
All $K$ robots begin at a base node $v_b \in \mathcal{V}$. 
Regarding reachability, we assume $G$ is strongly connected---but, not necessarily complete. 

Owing to unpredictable and/or uncertain hazards in the environment, a robot incurs a probability of destruction of $1 - \omega(v, v^\prime)$ when, beginning at node $v$, it attempts to traverse arc $(v, v^\prime) \in \mathcal{E}$ to visit node $v^\prime$.
Each outcome (survival or destruction) is an independent event. 
The arc survival probability map $\omega: \mathcal{E} \rightarrow (0, 1]$ is known, static over the course of the mission, and not necessarily symmetric (owing to e.g., [directional] air or water currents or bright sunlight or an adversary with limited range nearer one node than another).



Each node $v\in \mathcal{V}$ of the graph $G$ offers a reward $r(v) \in  \mathbb{R}_{\geq 0}$ to the robot team if visited by one or more robots over the course of the mission.
The reward $r(v)$ quantifies the utility gained by the team when a robot e.g.\ delivers a resource to node $v$, takes an image of node $v$ and transmits it back to the command center, or actuates some process (e.g.\ turns a valve) at node $v$. 
The total reward collected by the team is additive among nodes of the graph. 
Note, 
(1) even if a robot is destroyed after leaving a node, the harvested reward from that node is still accumulated by the team; and
(2) multiple visits to a node by the same or distinct robot(s) do not give marginal reward over a single visit. 
To collect rewards in this hazardous environment, the team of robots must plan a set\footnote{Since the robot team is homogenous, we consider the team trail plan as permutation-invariant and thus treat it as a set not a list.} of closed, directed trails $\mathcal{P}:=\{\rho_1, ..., \rho_K\}$ on the graph $G$ to follow.
A \emph{directed trail} \cite{clark1991first,graphtheory2} is an ordered list of nodes $\rho = (\rho[0], \rho[1], ..., \rho[\lvert \rho \rvert])$ where
(i) $\rho[i] \in \mathcal{V}$ is the $i$th node in the trail,  
(ii) an arc exists from each node in $\rho$ to the subsequent node, i.e., $(\rho[i-1], \rho[i])\in\mathcal{E}$ for $1 \leq i  \leq \lvert \rho \rvert$,
(iii) $\lvert \rho \rvert$ is the number of arcs traversed in the trail,
and
(iv) the arcs traversed in the trail are unique, i.e. each arc in the multiset $\{(\rho[i-1], \rho[i])\}_{i=1}^{\lvert \rho \rvert}$ has a multiplicity of one.
Note, unlike a path, the nodes in a trail are not necessarily distinct \cite{wilson1979introduction}.
A \emph{closed} trail begins and ends at the same node, i.e. $\rho [0]=\rho[\lvert \rho \rvert]$, which, here, $=v_b$.
Each trail $\rho_k$ belonging to the robot-team trail plan $\mathcal{P}$ constitutes a \emph{plan} because robot $k$ may be destroyed in the process of following $\rho_k$ and thus not \emph{actually} visit all nodes in $\rho_k$. A robot \emph{survives} the mission if it visits all nodes in its planned trail and returns to the base node (without getting destroyed).

We wish to design the robot-team trail plan $\mathcal{P}$ to maximize two objectives: 
(1) the expected team \underline{r}eward, $\mathbb{E}[R]$, and (2) the expected number of robots that \underline{s}urvive the mission, $\mathbb{E}[S]$. (Both $R$ and $S$ (i) are random variables owing to the stochasticity of robot survival while trail-following and (ii) depend on the robot-team trail plan $\mathcal{P}$ owing to different rewards among nodes and dangers among arcs.) 
That is, the BOTHE problem is:
\begin{equation}
	\max_{\mathcal{P}=\{\rho_1, ..., \rho_K\}} \left( \mathbb{E}[R(\mathcal{P})], \mathbb{E}[S(\mathcal{P})] \right)
	\label{eq:the_two_objs}
\end{equation}
when given the directed graph $G$ as a spatial abstraction of the environment, 
the homogenous team of $K$ mobile robots starting at the base node $v_b$,
the node reward map $r: \mathcal{V} \rightarrow \mathbb{R}_{\geq 0}$, and the arc survival probability map $\omega : \mathcal{E} \rightarrow (0, 1]$.

Because the bi-objective optimization problem in eqn.~\ref{eq:the_two_objs} presents a conflict between designing the robot-team trail plan to maximize the expected reward \emph{and} the number of surviving robots, we seek the \emph{Pareto-optimal set} \cite{pardalos2017non,branke2008multiobjective} of team-robot trail plans. 
The conflict is: 
(1) to maximize survival, a risk-averse robot team would not visit a dangerous region even if large rewards were contained therein, sacrificing reward for survival; 
(2) to maximize reward, a daring robot team would visit a dangerous region even if only small rewards were contained therein, sacrificing survival for reward. 
Hence, a utopian robot-team trail plan that simultaneously maximizes \emph{both} reward and survival objectives is unlikely to exist; the ultimate team trail plan selected for the mission must strike some tradeoff between reward and survival.
By definition, a \emph{Pareto-optimal} \cite{pardalos2017non,branke2008multiobjective} robot-team trail plan $\mathcal{P}^*$ cannot be altered to increase the survival objective $\mathbb{E}[S(\mathcal{P}^*)]$ without compromising (decreasing) the reward objective $\mathbb{E}[R(\mathcal{P}^*)]$---and vice versa. See Fig.~\ref{fig:pareto_optimal} and the formal definition below.
The Pareto-optimal set of team trail plans is valuable for presenting a \emph{portfolio} of team trail plans to a human decision-maker, who ultimately invokes their values---i.e., makes a team-reward vs.\  robot-survival tradeoff---by selecting a good team trail plan from the Pareto set.

\paragraph{Formal definition of Pareto-optimality and Pareto-front.}
Plan $\mathcal{P}^*$ belongs to the Pareto-optimal set of plans if and only if no other plan $\mathcal{P}^\prime$ \emph{Pareto-dominates} it. By definition, a plan $\mathcal{P}^\prime$ Pareto-dominates plan $\mathcal{P}^*$ if and only if both:
\begin{align}
	\left (\mathbb{E}[R(\mathcal{P}^\prime)] \geq \mathbb{E}[R(\mathcal{P}^*)]  \right) & \wedge \left( \mathbb{E}[S(\mathcal{P}^\prime)] \geq \mathbb{E}[S(\mathcal{P}^*)] \right) \\
	\left( \mathbb{E}[R(\mathcal{P}^\prime)] > \mathbb{E}[R(\mathcal{P}^*)] \right) & \vee \left( \mathbb{E}[S(\mathcal{P}^\prime)] > \mathbb{E}[S(\mathcal{P}^*)] \right).
\end{align}
So, the Pareto-dominating plan $\mathcal{P}^\prime$ is not worse than plan $\mathcal{P}^*$ for reward nor for survivability, and is better for one or both of them.
If a plan $\mathcal{P}^\prime$ were to Pareto-dominate another plan $\mathcal{P}^*$, one would objectively never choose plan $\mathcal{P}^*$ over plan $\mathcal{P}^\prime$, regardless of how they relatively value the two objectives. The \emph{Pareto front} is the set of objective vectors $\{(\mathbb{E}[R(\mathcal{P}^*)], \mathbb{E}[S(\mathcal{P}^*)])\}$ associated with the Pareto-optimal set of team trail plans $\{\mathcal{P}^*\}$.  

\paragraph{Comparison with TSOP.}
The BOTHE problem formulation follows the offline TSOP \cite{jorgensen2018team} with three modifications: we 
(i) omit the constraints that each robot survives above a threshold probability\footnote{i.e., we accept if one [uncrewed] robot bears much more risk of destruction than another during the mission.},
(ii) allow robots to follow trails instead of restricting to paths, as paths prevent robots from visiting a given node more than once and thus from e.g., harvesting reward from a leaf node having one in- and one out-degree involving the same node, 
and
(iii) have two objectives instead of one and seek the Pareto-optimal set of team trail plans.

\subsection{Probability distributions and expectations of $R(\mathcal{P})$ and $S(\mathcal{P})$}
\begin{figure}[h!]
    \centering
    	\includegraphics[width=0.6\textwidth]{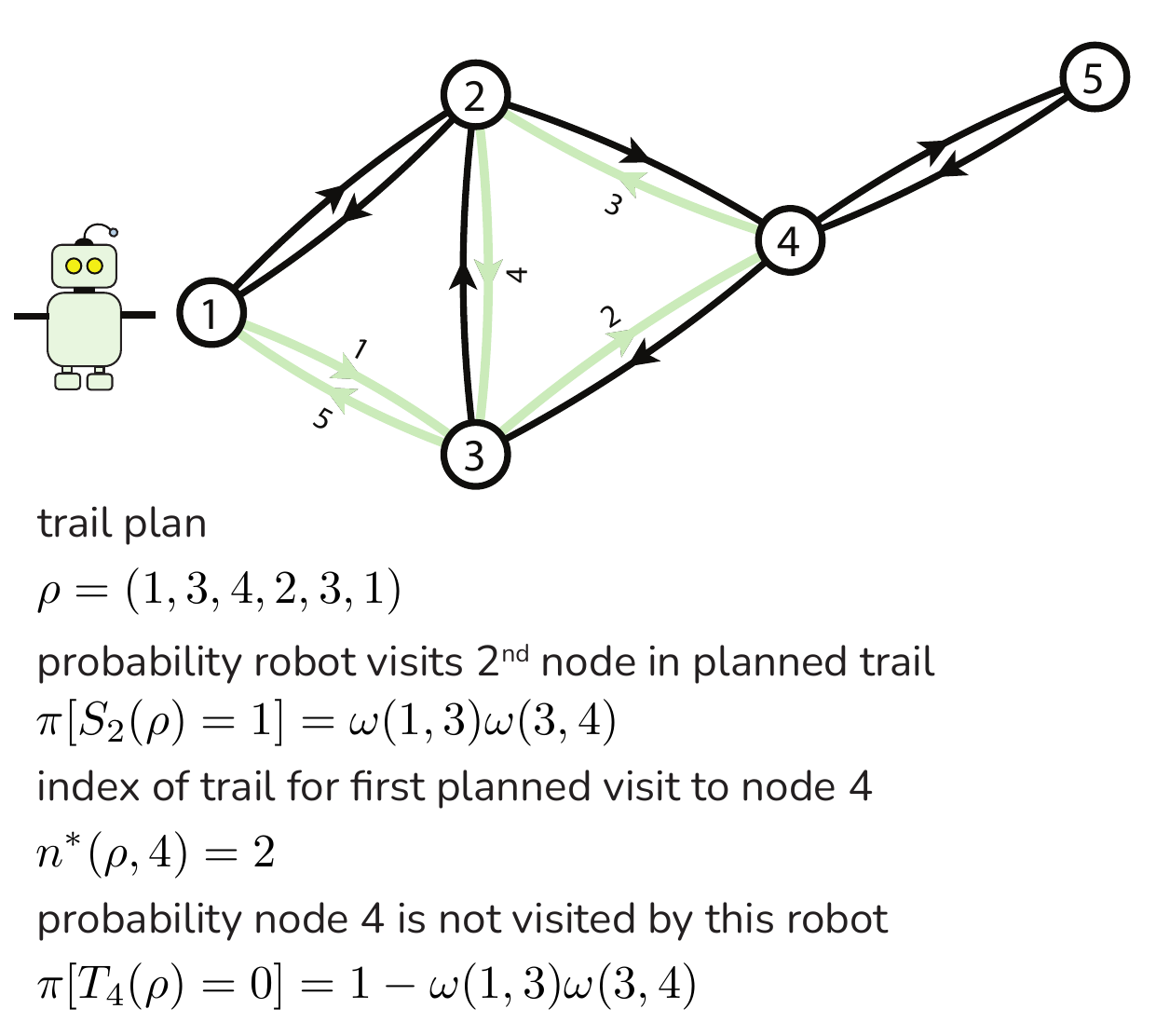}
    \caption{Illustrating notation for a particular robot's trail plan.} \label{fig:notation}
\end{figure}

We now write formulas for the two objectives---the expectations of the team reward $R(\mathcal{P})$ and of number of robots that survive the mission $S(\mathcal{P})$---as a function of the robot-team trail plan $\mathcal{P}$. 
These formulas follow from the directed graph $G$, arc survival probability map $\omega$, and node reward map $r$. Fig.~\ref{fig:notation} illustrates our notation.

\subsubsection{The survival of a single robot along its followed trail}
Central to computing both $\mathbb{E}[S(\mathcal{P})]$ and $\mathbb{E}[R(\mathcal{P})]$ is the probability that a robot survives to reach a given node in its planned trail.
Let $S_n(\rho)$ for $0 \leq n \leq \lvert \rho \rvert$ be the Bernoulli random variable that is one if a robot following trail $\rho$ survives to visit the $n$th node in the trail and zero otherwise. 
For the event of survival, the robot must survive traversal of \emph{all} of the first $n$ arcs in its trail to visit node $\rho[n]$. So, since the survival of a robot over each arc traversal attempt is an independent event, the probability that a robot following trail $\rho$ successfully visits node $\rho[n]$ is the product of the survival probabilities of the first $n$ arc-hops in the trail:
\begin{equation}
	\pi[S_n(\rho) = 1] = \prod_{i=1}^n \omega(\rho[i-1], \rho[i]) 
	= 1 - \pi[S_n(\rho) = 0]. \label{eq:pi_S_n}
\end{equation} 
The second equality follows because the complement of the event of survival is destruction.

\subsubsection{Expected number of robots that survive}
Now, we write a formula for the expected number of robots that survive the mission, $\mathbb{E}[S(\mathcal{P})]$. The event of survival of each robot is independent of the other robots.
Consequently, the number of robots that survive the mission, $S$, is the sum of the Bernoulli random variables indicating the survival of each robot over its planned trail:
\begin{equation}
	S(\mathcal{P}=\{\rho_1, ..., \rho_K\})=\sum_{k=1}^K S_{\lvert \rho_k \rvert}(\rho_k). \label{eq:R_sum}
\end{equation}
Thus, $S$ follows the Poisson-Binomial distribution \cite{tang2023poisson}.
Seen from eqn.~\ref{eq:R_sum} and the linearity of the expectation operator, the expected number of robots that survive the mission is:
\begin{equation}
	\mathbb{E}[S(\mathcal{P}=\{\rho_1, ..., \rho_K\})]=\sum_{k=1}^K \mathbb{E}[S_{\lvert \rho_k \rvert}(\rho_k)] = \sum_{k=1}^K  \pi[S_{\lvert \rho_k \rvert}(\rho_k) = 1] \label{eq:formula_obj2}
\end{equation} with $\pi[S_{\lvert \rho_k \rvert}(\rho_k) = 1]$ given in eqn.~\ref{eq:pi_S_n}.

\subsubsection{Expected team reward}
Now, we write a formula for the expected team reward, $\mathbb{E}[R(\mathcal{P})]$. 

First, we calculate the probability that a robot following a given trail $\rho$ does not visit a given node $v\in \mathcal{V}$.
Let the Bernoulli random variable $T_v(\rho)$ be one if a robot following trail $\rho$ visits node $v$ and zero if it does not.
If node $v$ is not in the planned trail $\rho$, $T_v(\rho)=0$ with certainty. 
Now, suppose node $v$ is in the planned trail $\rho$. Let $n^*$ be the index in the trail where the robot plans its first visit to node $v$:
\begin{equation}
n^*(\rho, v) = \min_{
	\substack{n \in \{0, ..., \lvert \rho \rvert\} \\ \rho[n] = v}
} n.
\end{equation}
Then, $T_v(\rho)$ is equal to the random variable $S_{n^*}(\rho)$ because the robot visits node $v$ if and only if it survives its first $n^*$ arc-hops to first land on node $v$. 
So, the probability node $v$ is not visited by a robot following trail $\rho$ is:
\begin{equation}
	\pi[T_v(\rho) = 0] = 
	\begin{cases}
		1 & v\notin \rho\\
		 \pi [S_{n^*(\rho, v)}(\rho)=0 ] & v \in \rho
	\end{cases}
	 \label{eq:pi_T_v}
\end{equation}
with $\pi[S_{n^*}(\rho)=0]$ calculated using eqn.~\ref{eq:pi_S_n}.

Second, we calculate the probability that a given node $v\in\mathcal{V}$ is visited by one or more robots on a team following trail plans $\mathcal{P}=\{\rho_1, ..., \rho_K\}$; only then is the reward offered by that node, $r(v)$, accumulated by the team.
Let $T_v(\mathcal{P})$ be the Bernoulli random variable that is one if one or more robots on the team with trail plans $\mathcal{P}$ visit node $v$ and zero otherwise (i.e.\ if and only if zero of the robots visit $v$).
The complement of the event $T_v(\mathcal{P})=1$ is that all $K$ robots do not visit node $v$ (independent events), so:
\begin{equation}
	\pi [T_v( \{\rho_1, ..., \rho_K\} ) = 1] = 
	1 - \prod_{k=1}^K \pi[T_v(\rho_k)=0].
	\label{eq:pi_T_v_all}
\end{equation} 
with $\pi[T_v(\rho) = 0]$ in eqn.~\ref{eq:pi_T_v}.

Finally, the total team reward collected by the robot-team following trail plan $\{\rho_1, ..., \rho_K\}$ is the sum of the rewards given to the team by each node:
\begin{equation}
R(\{\rho_1,...,\rho_K\}) = \sum_{v\in\mathcal{V}} r(v)  T_v(\{\rho_1, ..., \rho_K\}),
\end{equation} where the reward from node $v$, $r(v)$, is accumulated if and only if node $v$ is visited by one or more robots (i.e. iff $T_v(\mathcal{P})=1$).
The expected reward accumulated over the mission by a robot-team following trail plans $\{\rho_1, ..., \rho_K\}$ is:
\begin{equation}
	\mathbb{E}[R(\{\rho_1,...,\rho_K\})]= \sum_{v\in\mathcal{V}} r(v) \pi[T_v(\{\rho_1, ..., \rho_K\}) = 1] \label{eq:formula_obj1}
\end{equation}
with $ \pi[T_v(\{\rho_1, ..., \rho_K\}) = 1]$ given in eqn.~\ref{eq:pi_T_v_all}.


\section{Bi-objective ant colony optimization to search for the Pareto-optimal robot-team trail plans}
To efficiently search for the Pareto-optimal set of robot-team trail plans defined by eqn.~\ref{eq:the_two_objs}, we employ bi-objective (BO) ant colony optimization (ACO) \cite{iredi2001bi}. ACO \cite{dorigo2006ant,bonabeau1999swarm,blum2005ant} is a meta-heuristic for combinatorial optimization problems framed as a search for an optimal path (or trail) on a graph. As a swarm intelligence method \cite{bonabeau1999swarm}, 
ACO is inspired by the behavior of ants efficiently foraging for food \cite{bonabeau2000inspiration}. See Box~\ref{box:ants}. 
As precedent, variants of the single-robot- and team-orienteering problem have been efficiently and effectively solved by ACO \cite{ke2008ants,chen2015multiobjective,verbeeck2017time,sohrabi2021acs,chen2022environment,montemanni2011enhanced} (as well as other [meta-]heuristics \cite{gavalas2014survey,dang2013effective,chao1996fast,butt1994heuristic}), but not the TSOP.

\begin{mybox}[label=box:ants, breakable]{Pheromone trail laying and following by foraging ants}
Largely (but not exclusively \cite{evison2008combined,czaczkes2015trail,robinson2005no}) via laying and following pheromone trails \cite{czaczkes2015trail}, 
some species of foraging ant colonies advantageously \cite{deneubourg1983probabilistic} can collectively select 
(i) the shortest path from the nest to a food source \cite{goss1989self}
and
(ii) the highest quality food source among multiple options equidistant from the nest \cite{beckers1993modulation}. 

Pheromone, a relatively volatile chemical \cite{david2009trail}, serves as an olfactory cue for ants \cite{knaden2016sensory}, allowing for \emph{indirect} communication between ants in the colony. 
On the way back to the nest from a food source, ants deposit pheromone on the ground---an amount modulated by the quality of the food source \cite{beckers1993modulation}.
Shorter paths to higher quality food sources accumulate pheromone more quickly.
Since ants are recruited to and follow pheromone \cite{beckers1993modulation,czaczkes2015trail}, these paths recruit more ants and get reinforced. Via this positive feedback, the colony can collectively select the shortest path or highest-quality food source \cite{jackson2006communication,czaczkes2015trail,bonabeau1999swarm}.
The pheromone trails laid by the colony form a \emph{collective memory}---a local guide for ants to high-quality food sources and short paths to them \cite{jackson2006communication}.
Some species of ants can deposit multiple species of pheromone (from different glands) with e.g.\ different longevities \cite{czaczkes2015trail}, allowing for more complex indirect communication \cite{jackson2006communication,robinson2005no}.

As negative feedback mechanisms, 
(i) pheromone evaporates over time \cite{jackson2006communication,david2009trail,van2011temperature}, allowing the ant colony to ``forget'' trails to exhausted food sources,
and 
(ii) ants deposit less pheromone on trails a) with already-high pheromone concentrations \cite{czaczkes2013ant} or b) leading to food sources already occupied by their nestmates \cite{wendt2020negative}.

Ants select among pheromone trails with some degree of stochasticity \cite{deneubourg1990self}, which is beneficial for 
(i) continual exploration to find even shorter paths and even higher-quality food sources, 
(ii) exploiting multiple food sources in parallel, 
and 
(iii) plasticity in a dynamic environment \cite{deneubourg1983probabilistic,shiraishi2019diverse,deneubourg1986random,dussutour2009noise,edelstein1995trail}.

The combination of positive feedback, negative feedback, and randomness in ants' pheromone-laying and -following can give rise to complex collective behavior despite simple interactions among decentralized individuals \cite{bonabeau1997self,bonabeau1999swarm,goss1989self,jackson2006communication,edelstein1995trail,watmough1995modelling}.
\end{mybox}

In bi-objective ACO, we simulate a heterogenous colony of artificial ants walking on the graph $G$---under a loose analogy, foraging for food---over many iterations. 
At each iteration, each worker ant stochastically constructs a robot-team trail plan $\{\rho_1, ..., \rho_K\}$ robot-by-robot, arc-by-arc, biased by (i) the amounts of two species of pheromone on the arcs encoding the colony's past experiences and (ii) two heuristics that score the greedy, \emph{a priori} appeal of each arc for each objective. 
As a division of labor, each worker ant specializes by searching for team trail plans belonging to a different region of the Pareto-front.
Then, ants that found the Pareto-optimal team trail plans over that iteration deposit pheromone of each species on the arcs involved, proportionally to the value of the objective achieved by that plan.
An elitist ant \cite{dorigo1996ant} maintains a set of global (i.e, over all iterations) Pareto-optimal team trail plans and also deposits pheromone on arcs.
Finally, to prevent stagnation and promote continual exploration, a fraction of the pheromone evaporates. After many iterations, the ACO algorithm returns the [approximate\footnote{The ACO meta-heuristic is not guaranteed to find all Pareto-optimal solutions nor neglect to include a Pareto-dominated solution.}] Pareto-optimal set of robot-team trail plans maintained by the elitist ant.

\subsection{The heterogenous artificial colony of ants, pheromone, and heuristics}
Our heterogenous artificial ant colony consists of (i) $N_{\text{ants}}$ worker ants and (ii) an elitist ant.
Worker ant $i\in\{1, ..., N_{\text{ants}}\}$ in the colony is assigned a parameter $\lambda_i := (i-1) / (N_{\text{ants}}-1)$ dictating its balance of the two objectives when searching for Pareto-optimal team trail plans.
A $\lambda$ closer to zero (one) implies the ant prioritizes maximizing the expected reward $\mathbb{E}[R]$ (robot survivals $\mathbb{E}[S]$). 
So, different ants seek team trail plans belonging to different regions of the Pareto front.
The elitist ant maintains the global-Pareto-optimal set of team trail plans.

Each arc $(v, v^\prime)\in\mathcal{E}$ of the graph is associated with 
(i) amounts of two distinct species of pheromone, $\tau_R(v, v^\prime)$ and $\tau_S(v, v^\prime)$, and 
(ii) heuristic scores $\eta_R(v, v^\prime)$ and $\eta_S(v, v^\prime)$.
Both $\tau_{R,S}(v, v^\prime)>0$ and $\eta_{R,S}(v, v^\prime)>0$ score the promise of arc $(v, v^\prime)$ for belonging to Pareto-optimal team trail plans that maximize $\mathbb{E}[R]$ and $\mathbb{E}[S]$, respectively, and guide worker ants' construction of robot-team trail plans.
The pheromone is learned, reflecting the past collective experience/memory of the ant colony. 
Due to deposition and evaporation, the pheromone is dynamic over iterations.
By contrast, the heuristic is static and scores the \emph{a priori}, greedy/myopic appeal of each arc to accelerate the convergence of ACO.
%
%
%

\subsection{Constructing robot-team trail plans}
Each iteration, every worker ant stochastically constructs a team trail plan $\mathcal{P}=\{\rho_1, ..., \rho_K\}$ by sequentially allocating trails to the robots and [conceptually] following the closed trail the ant designs for each robot. Computing the objectives achieved under each plan via eqns.~\ref{eq:formula_obj2} and \ref{eq:formula_obj1} gives data $\{ (\mathcal{P}_i, \mathbb{E}[R(\mathcal{P}_i)], \mathbb{E}[S(\mathcal{P}_i)])\}_{i=1}^{N_{\text{ants}}}$ used to deposit pheromone (worker ant) and update the global Pareto-optimal set (elitist ant). 

A robot trail is constructed by iteratively applying a stochastic, partial-trail extension rule until the closed trail is complete. Suppose an ant with objective-balancing parameter $\lambda$ is constructing the closed trail for robot $k$, $\rho_k$, and currently resides at node $v=\rho_k[i]$.
Namely, the ant has selected (i) the trails for the previous $k-1$ robots, $(\rho_1, ..., \rho_{k-1})$, and (ii) an incomplete/partial trail for robot $k$, $\tilde{\rho_k}=(v_b, \rho_k[1], ..., \rho_k[i]=v)$, giving its first $i$ arc-hops.
The probability of next hopping to node $v^\prime$ across arc $(v, v^\prime)\in\mathcal{E}$ not yet traversed in the partial trail $\tilde{\rho_k}$ is:
 \begin{equation}
	\pi(v^\prime \mid \rho_1, ..., \rho_{k-1}, \tilde{\rho_k}) \propto 
		 \left[\tau_R(v, v^\prime) \eta_R(v, v^\prime; \rho_1, ..., \rho_{k-1},\tilde{\rho_k}) \right]^{1-\lambda} \left[ \tau_S(v, v^\prime) \eta_S(v, v^\prime) \right]^\lambda.
	 \label{eq:prob_x_y}
\end{equation}
The partial trail is more likely to be extended with $v^\prime=:\rho_k[i+1]$ if arc $(v, v^\prime)$ has more pheromone $\tau_{R, S}(v, v^\prime)$ and/or greedy heuristic appeal $\eta_{R, S}(v, v^\prime)$---with more or less emphasis on the reward or survivability pheromone/heuristic depending on the ant's $\lambda$.
Iteratively extending the partial trail using eqn.~\ref{eq:prob_x_y}, the ant completes the trail for robot $k$ after it traverses the self-loop of the base node, $(v_b, v_b)$. Then, the ant begins trail construction for robot $k+1$ if $k<K$ or completes its team trail plan $\mathcal{P}$ if $k=K$.


\paragraph{Heuristics.} 
For the survivability objective, we score the desirability of arc $(v, v^\prime)$ with the probability of the robot surviving traversal of that arc, $\eta_S(v, v^\prime):=\omega(v, v^\prime)$---myopic because it does not consider survivability of arcs later in the trail. 
For the reward objective, we greedily score the desirability of arc $(v, v^\prime)$ with the expected marginal reward the team receives by robot $k$ visiting node $v^\prime$ next, which is $r(v^\prime)$ if 
(i) none of the previous $k-1$ robots visit node $v^\prime$, 
(ii) node $v^\prime$ is not planned to be visited earlier in the trail of robot $k$, and
(iii) robot $k$ survives its hop to node $v^\prime$
and zero otherwise:
\begin{equation}
	\eta_R(v, v^\prime; \rho_1, ..., \rho_{k-1}, \tilde{\rho_k}) :=  
	 \pi[ T_{v^\prime}(\{\rho_1, ..., \rho_{k-1}\}) = 0)] \mathcal{I}[v^\prime \notin \tilde{\rho_k}] \omega(v, v^\prime) r(v^\prime ) ,
\end{equation}
with $\mathcal{I}(\cdot)$ the indicator function---myopic because it does not account for rewards robot $k$ could collect further along the trail. 
Note, to prevent either heuristic from being exactly zero (resulting in \emph{never} selecting that arc), we add a small number $\epsilon$ to each heuristic.

\subsection{Pheromone update}
At the end of each iteration, we update the pheromone maps $\tau_R$ and $\tau_S$ to capture the experience of the ants in finding Pareto-optimal robot-team trail plans, better-guide the ants' trail-building in the next iteration, and prevent stagnation (premature convergence).

The pheromone update rule is:
\begin{equation}
	\tau_{R, S}(v, v^\prime) \leftarrow (1-\rho) \tau_{R,S}(v, v^\prime)  + \Delta \tau_{R,S}(v, v^\prime) \text{ for } (v, v^\prime) \in \mathcal{E}, \label{eq:tau_update}
\end{equation}
with $\rho \in (0, 1)$ the evaporation rate (a hyperparameter) and $\Delta \tau_{R,S}(v, v^\prime)$ the amount of new pheromone deposited on arc $(v, v^\prime)$ by the ants.

Evaporation, accomplished by the first term in eqn.~\ref{eq:tau_update}, removes a fraction of the pheromone on every arc. This negative feedback mechanism prevents premature convergence to suboptimal trails and encourages continual exploration.

Deposition, accomplished by the second term in eqn.~\ref{eq:tau_update}, constitutes indirect communication to ants in future iterations about which arcs tend to belong to Pareto-optimal team trail plans and the objectives achieved under those plans.
First, the ants \emph{collaborate} by 
(i) among the worker ants, comparing the solutions constructed \emph{this} iteration to obtain the \emph{iteration}-Pareto-optimal set of plans
and 
(ii) worker ants sharing the solutions constructed this iteration with the elitist ant, who then updates the \emph{global}-Pareto-optimal set.
Second, the worker ants with an iteration-Pareto-optimal team trail plan execute their constructed plan while depositing pheromone on the arcs.
Third, the elitist ant executes all global-Pareto-optimal team trail plans while depositing pheromone on the arcs. Each ant deposits both reward and survival pheromone in proportion to the reward and survival objectives, respectively, achieved under the plan they are following. (The objective is analogous with food quality). 
Specifically, amalgamating the iteration- and global-Pareto-optimal team trail plans into a multiset $\{(\mathcal{P}_p^*, \mathbb{E}[R(\mathcal{P}_p^*)], \mathbb{E}[R(\mathcal{P}_p^*)])\}_{p=1}^P$, arc $(v, v^\prime)$ receives pheromone:
\begin{equation}
	 \Delta \tau_{R,S}(v, v^\prime) := 
	\frac{1}{P} \sum_{p=1}^{P} \mathbb{E}[R(\mathcal{P}^*_p), S(\mathcal{P}^*_p)] 
	\sum_{k=1}^K 
	\sum_{i=1}^{\lvert \rho_k^{(p)}\rvert}
	\mathcal{I} \left[ 
		(v, v^\prime)=(\rho_k^{(p)}[i-1], \rho_k^{(p)}[i])
	\right].
\end{equation}
with $\mathcal{I}(\cdot)$ the indicator function.
The sums are over, left-to-right, Pareto-optimal plans, robot trails in those plans, and arcs in those robot trails. 
The latter two sums count the number of times the arc $(v, v^\prime)$ appears in $\mathcal{P}^*_p$.
By construction, arcs that receive the most reward (survival) pheromone frequently belong to trails in the Pareto-optimal team trail plans with high reward (survival).

We initialize the pheromone on all arcs with $\tau_{R,S}(v, v\prime)=1$, to allow the heuristic to completely guide the first iteration. 


\subsection{Area indicator for Pareto-set quality}
From iteration-to-iteration, we measure the quality of the global [approximate] set of Pareto-optimal robot-team trail plans tracked by the elitist ant using an area indicator \cite{cao2015using,guerreiro2020hypervolume}---loosely, the area in objective space enclosed between the origin and the [approximated] Pareto-front. Formally, the quality $q$ of a Pareto set $\{\mathcal{P}^*_1, ...,\mathcal{P}^*_P\}$ is the area of the union of rectangles in 2D objective space:
\begin{equation}
	q(\{\mathcal{P}^*_1, ...,\mathcal{P}^*_{P}\}):=
	\Big \lvert 
		\bigcup_{p=1}^P \{ o \in \mathbb{R}^2 : o \geq 0 \wedge  o \leq (\mathbb{E}[R(\mathcal{P}^*_p)], \mathbb{E}[S(\mathcal{P}^*_p)]) \} 
	\Big \rvert \label{eq:q}
\end{equation}
illustrated with the shaded yellow area in Fig.~\ref{fig:pareto_optimal}.

\section{Results}
We now implement bi-objective ant colony optimization to search for Pareto-optimal robot-team trail plans for an instance of bi-objective team orienteering in a hazardous environment. 
We visualize the pheromone trail and some of the Pareto-optimal solutions to gain intuition. We also conduct an ablation study to quantify the importance of the pheromone and heuristics for guiding the search.
Our Julia code to reproduce our results and modify and build upon with more complexity is available at \url{github.com/SimonEnsemble/BO_ACO_TOHE}.

\subsection{Problem setup} 
A team of $K=3$ mobile robots are assigned an information-gathering mission in the San Diego Museum of Art. 
We selected this art museum for (i) its rich connectivity between galleries, giving an interesting example, and (ii) its small size, allowing us to visualize, interpret, and intuit the solution to a BOTOHE problem on it.
We spatially model the art museum as a directed graph $G=(\mathcal{V}, \mathcal{E})$.
The set of 27 nodes $\mathcal{V}$ represents the 23 art galleries (rooms), the outside entrance to the building (base node $v_b$), the main entrance rooms on the first and second floors, and the stairway.
The set of arcs $\mathcal{E}$ represents direct passages/doorways between the rooms. 

Suppose traversing the art museum is hazardous for the robots, owing to
(i) adversarial security guards that (a) seek to prevent the robots from imaging the art and (b) possess the ability attack and/or capture the robots, and
(ii) obstacles that the robots could (a) crash into or (b) become entangled in.
To model risks of destruction or capture, we assign survival probabilities $\omega(i,j)$ for the arc(s)
(i) traversing the staircase of 0.8,
(ii) inside and in/out of the main entrance room of 0.9,
(iii) on the right side of the first floor of 0.97,
(iv) on the left side of the first floor of 0.95,
and
(v) on the second floor of 0.9.

Suppose, when a robot visits an art gallery in the museum, it images the art there and transmits this image back to the command center, providing utility to the command center. 
The utility of each image to the command center is scored by
the node reward map $r$ assigning rewards of
(i) 2/3 for large galleries,
(ii) 1/3 for medium-sized galleries,
(iii) 1 for small galleries that we suppose contain the most valuable art, and
(iv) 1/10 for five galleries that are in corners of the museum or behind the stairway.

The two objectives of the command center are to plan the trails of the robots in the art museum to maximize the (1) expected reward, via robots visiting art galleries, imaging them, then transmitting the images back to the command center, and (2) expected number of robots that return from the mission. 

\begin{figure}[h!]
    \centering
    	\includegraphics[width=0.6\textwidth]{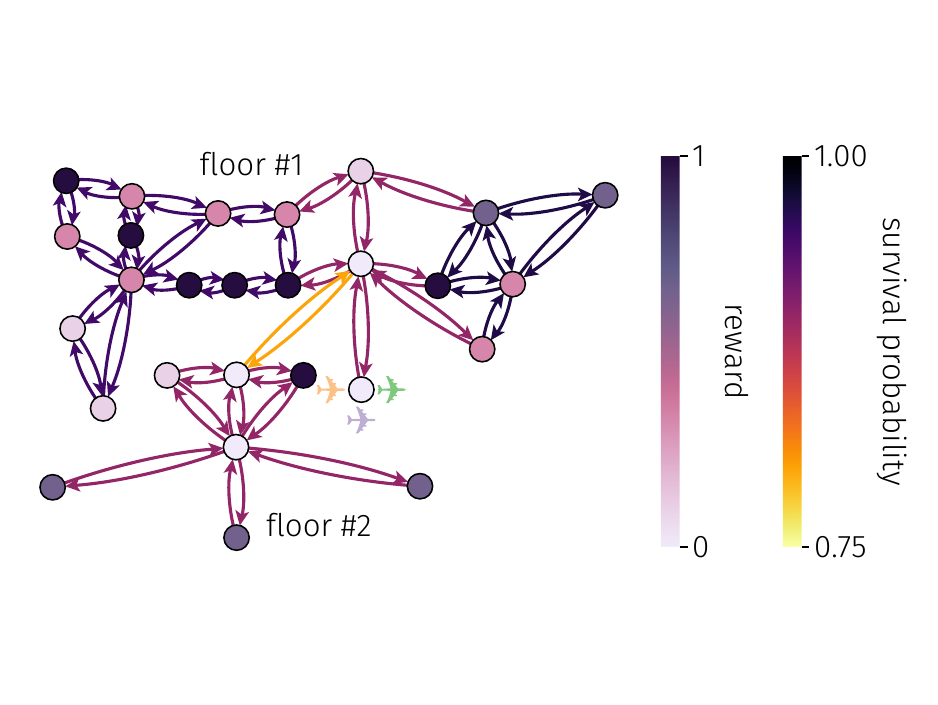}
    \caption{Our TOHE problem instance. The directed graph represents the two-floor San Diego art museum with distinct rooms (nodes) connected by doorways or a stairway (arcs). The arcs are colored according to robot survival probabilities. The nodes are colored according to rewards offered to the team when a robot visits. The three robots (planes) initially reside at the base node. 
    } \label{fig:ex_setup}
\end{figure}

We visualize this BOTOHE problem instance in Fig.~\ref{fig:ex_setup}. 
The topology of the directed graph is shown, with a layout reflecting the spatial location of the rooms in the San Diego Museum of Art. The nodes on the first and second floor are grouped together.
The base node is marked by the three robots near it.
The arc survival probability map $\omega$ is visualized by the color assigned to each arc.
The stairway is the most dangerous arc.
The node reward map $r$ is visualized by the color assigned to each node. 


\subsection{Computational results}
We now employ our bi-objective ant optimization algorithm to search for the Pareto-optimal robot-team trail plans for our problem instance.
We use a colony of $N_{ants}=100$ artificial ants, a pheromone evaporation rate of $\rho=0.04$, and 10\,000 iterations. We initialize the pheromone maps with one unit of pheromone on each arc.
The runtime is $\sim$5\,min on an Apple M1 machine.

\begin{figure}[h!]
    \centering
    \includegraphics[width=0.4\textwidth]{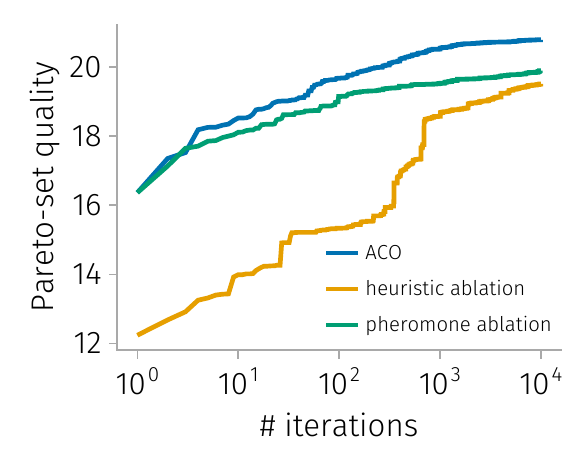}
    \caption{Progress of BO-ACO on our TOHE problem instance and comparison with baselines. 
    The quality of the Pareto-set of robot team trail plans, measured via the area indicator, is shown as a function of the number of iterations of ACO---for ordinary ACO, ACO without heuristics, and ACO without pheromone.
    } \label{fig:aco_progress}
\end{figure}

Fig.~\ref{fig:aco_progress} shows the quality (area indicator) of the Pareto-set ($q$ in eqn.~\ref{eq:q}) over iterations. As the ACO algorithm progresses, the quality of the Pareto-set improves, but with diminishing returns. The saturation of the progress over iterations indicates satisfactory convergence.

ACO searches for Pareto-optimal robot-team trail plans,
guided by (i) the static heuristics $\eta_{R,S}$ that greedily score the appeal of traversing a given arc and (ii) the dynamic pheromone maps $\tau_{R,S}$ that encapsulate the memory of the ant colony over previous iterations. 
Next, we quantify how each of these components are contributing to the effectiveness of ACO by ablating each. 
First, we run ACO where heuristics are not used to bias the ants towards \emph{a priori} promising arcs by setting $\eta_{R,S}(v, v^\prime)=1$ for all arcs $(v, v^\prime)\in \mathcal{E}$. Second, we run ACO where pheromone is not used to bias the ants towards promising arcs based on the history of the colony's search by setting $\tau_{R,S}(v, v^\prime)=1$ for all arcs $(v, v^\prime)\in \mathcal{E}$.
Fig.~\ref{fig:aco_progress} compares the search efficiency of these ablated runs with ordinary ACO (where both heuristics and pheromone are used).
Compared to ordinary ACO, the search efficiency diminishes for each ablation study. 
Thus, both heuristic and pheromone contribute to the search efficiency.
However, the search efficiency drops more dramatically when the heuristic is ablated. 
Thus, the heuristic is more helpful than the pheromone in terms of finding a quality Pareto-set in few iterations. 

\begin{figure}[h!]
    \centering
    \begin{subfigure}[b]{0.59\textwidth}
    	\includegraphics[width=\textwidth]{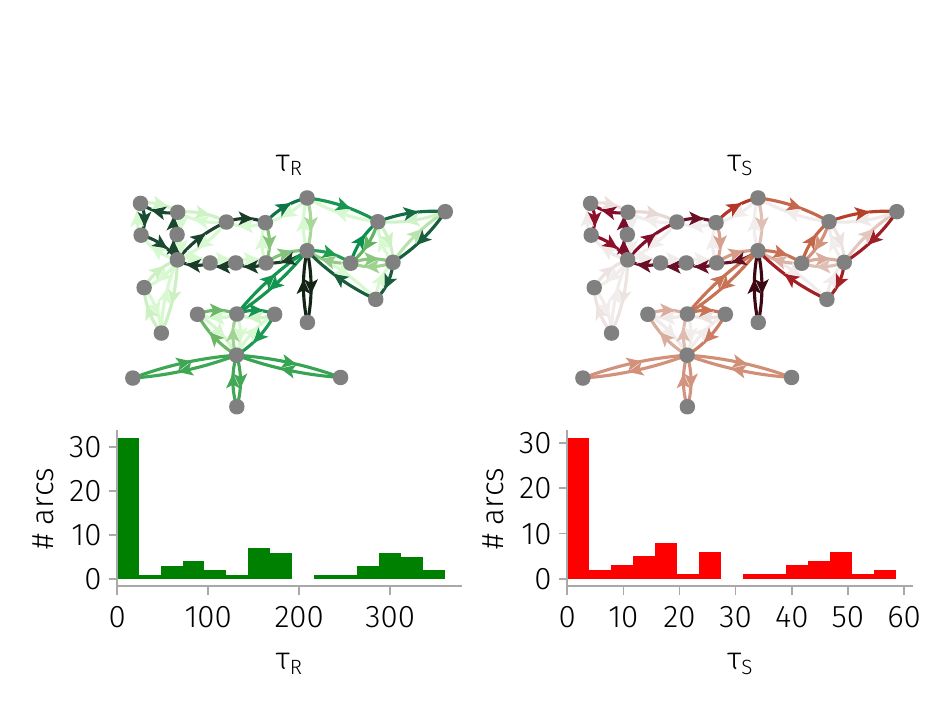}
	\caption{Pheromone trails} \label{fig:pheromone}
    \end{subfigure}
        \begin{subfigure}[b]{\textwidth}
    	\includegraphics[width=\textwidth]{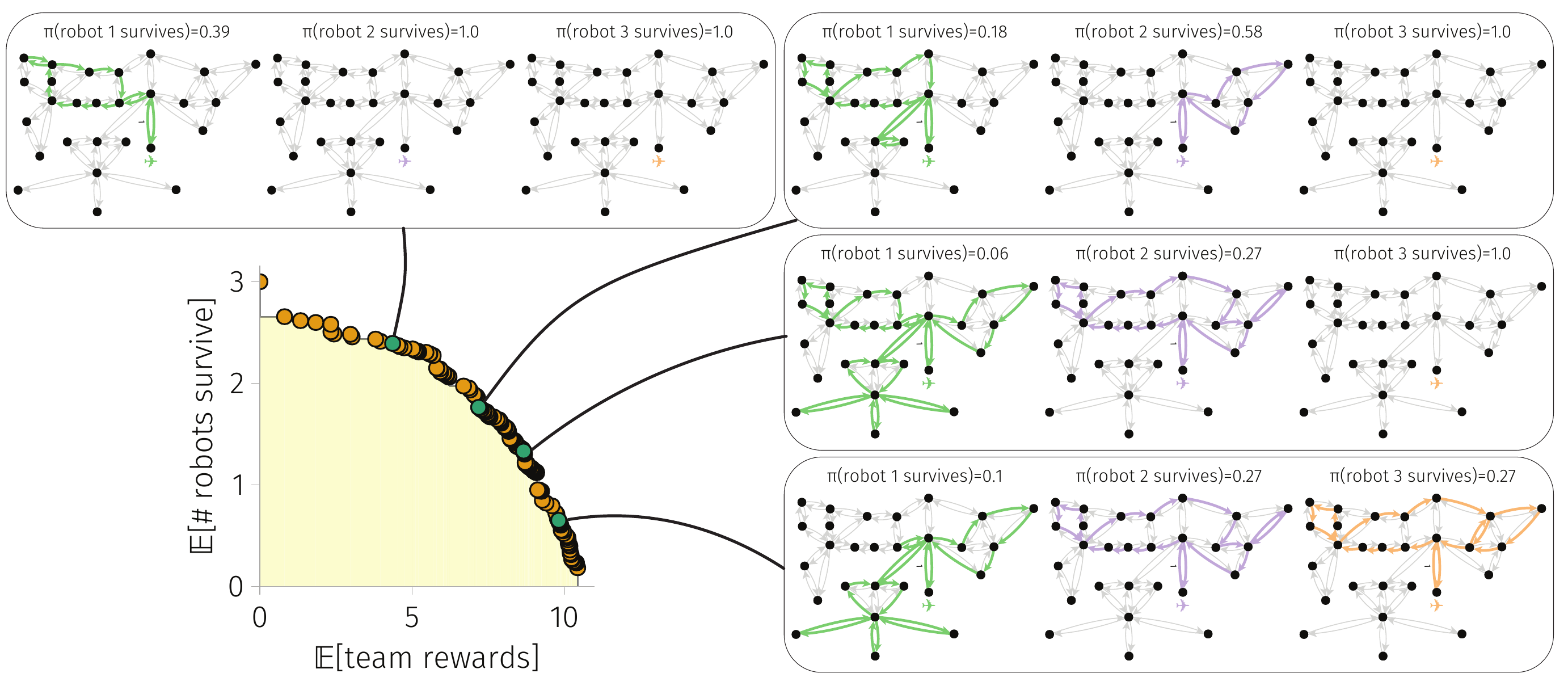}
	\caption{Pareto-optimal solutions} \label{fig:pareto_front}
    \end{subfigure}
    \caption{Analysis of the BO-ACO solution to our TOHE problem instance. 
    (a) The pheromone trails and distribution of the amount of pheromone on the arcs at the end of the ACO algorithm. 
    (b) The [approximate] Pareto-front of robot-team team trail plans at the end of the ACO algorithm with four select plans shown. The indicator of the Pareto-set quality is the area of the highlighted, yellow region.
    }
\end{figure}

At the end of the ACO algorithm, Fig.~\ref{fig:pheromone} visualizes the pheromone maps $\tau_{R}$ and $\tau_S$.
 Note the clock-wise vs.\ counter-close-wise preferences in some cycles of high pheromone to pursue high-reward nodes earlier in the trails to maximize expected reward.

Fig.~\ref{fig:pareto_front} shows the [approximate] Pareto-front of 167 solutions found by BO-ACO. 
Walking down the Pareto-front trades larger expected team reward for a smaller expected number of robots that return from the mission. The key motivation for treating the \emph{bi-objective} TOHE problem is to present this Pareto-front to a human decision-maker who ultimately chooses a robot-team trail plan that balances reward with robot survival.

Finally, Fig.~\ref{fig:pareto_front} shows four Pareto-optimal trail plans belonging to different regions of the Pareto-front. 
As we move down the front, plans offer larger expected reward but lower robot survival.
In the first plan, only a single robot enters the museum to image galleries offering the highest rewards on the left side of the first floor---avoiding the more dangerous right side of the second floor and the very dangerous staircase to the second floor. 
In the second plan, two robots are utilized: the first plans to cover much of the left side of the first floor, while the second plans to cover much of the right side of the first floor.
In the third plan, also utilizing two robots, the first robot plans to traverse most of the first floor and the second floor later. The second robot traverses much of the first floor. This redundancy in first-floor coverage builds robustness into the plans: even if the first robot gets destroyed early on, the second robot can still image most of the galleries on the first floor.
Finally, the fourth plan uses all three robots with much redundancy to achieve high expected reward. Still, the robots avoid taking the risk to enter the bottom left corner of the first floor, whose galleries offer only a small reward.

\section{Discussion}
Many applications for mobile robot teams---ranging from information-gathering, resource-delivery, chemical plume source localization, forest fire fighting, to resource delivery---involve traversing environments with hazards e.g., corrosive chemicals, attacking adversaries, obstacles, or rough terrain/seas. 
For each application, the robots must coordinate their trails for the team-level objective in a risk-aware manner: select the subset of locations to visit, assign the safest trails to the robots, balance reward and risk to the robots, and build redundancy into the plans to make the objective robust to the failure of robots on the team. 

Heavily inspired by the Team Surviving Orienteers Problem \cite{jorgensen2018team,jorgensen2017matroid,jorgensen2024matroid}, we posed the bi-objective team orienteering in a hazardous environment (BOTHE) problem. 
By finding the Pareto-optimal set of robot-team trail plans, we can present them to a human decision-maker who ultimately chooses the robot trail plans for the mission according to how he or she values the expected reward collected by the robot team compared with the expected number of robots that survive the dangerous mission.

We employed bi-objective ant colony optimization to search for the Pareto-optimal team trail plans. Despite lacking theoretical guarantees to find the Pareto-optimal set, ACO was effective and can be readily adapted to handle extensions to the TOHE problem.

\paragraph{Future work.}
For the BOTOHE we have posed, we wish to (i) tackle the online version with ACO, where the robots adapt their planned trails during the mission, in response to observed failures of robots, (ii) devise local search methods to improve the robot trails the ants found, accelerating the convergence of ACO \cite{dorigo2006ant}, (iii) benchmark non-sequential methods to allocate trails to robots with ACO \cite{ke2008ants}, and (iv) employ multi-colony ant optimization \cite{iredi2001bi}.

Interesting and practical extensions of robot-team orienteering in adversarial/hazardous environments abstracted a graphs include treating (some of these ideas from Ref.~\cite{jorgensen2018team}): 
(i) a heterogenous team of robots with different (a) capabilities to harvest rewards from nodes and (b) survival probabilities for each arc traversal owing to e.g.\ stealth;
(ii) more complicated reward structures, e.g., time-dependent, stochastic, non-additive (correlated \cite{yu2014correlated}), multi-category, or multi-visit rewards;
(iii) fuel/battery constraints of the robots and nodes representing refueling, recharging, or battery-switching stations \cite{asghar2023risk,khuller2011fill,liao2016electric,yu2019coverage}; 
(iv) constraints on the rewards a robot can harvest e.g.\ for resource delivery applications where each robot holds limited cargo capacity \cite{coelho2014thirty};
(v) non-binary surviving states of the robots due to various levels of damage;
(vi) non-independent events of robots surviving arc-traversals;
(vii) risk metrics different from the expected value \cite{majumdar2020should}.

Another interesting direction is to learn/modify the survival probabilities associated with the edges of the graph from data over repeated missions (an inverse problem \cite{burton1992instance}). 
Specifically, suppose we are uncertain about the survival probability $\omega(i, j)$ of each arc $(i,j)$. Within a Bayesian inference framework, we may impose a prior distribution on each $\omega(i,j)$. Then, when a robot survives or gets destroyed during a mission, we update the prior distribution. 
The trail-planning of the robots over sequential missions may then balance (a) exploitation to harvest the most reward and take what appear to be, under uncertainty, the safest trails and (b) exploration to find even safer trails.

Finally, instead of abstracting the environment as a graph, one could path-plan for robots in a continuous space with obstacles. 

\section*{Acknowledgements} CMS acknowledges the Office of Naval Research Summer Faculty Research Program.

\bibliography{refs}

\begin{thebibliography}{10}

\bibitem{leutenegger2016flying}
Stefan Leutenegger, Christoph H{\"u}rzeler, Amanda~K Stowers, Kostas Alexis,
  Markus~W Achtelik, David Lentink, Paul~Y Oh, and Roland Siegwart.
\newblock Flying robots.
\newblock {\em Springer Handbook of Robotics}, pages 623--670, 2016.

\bibitem{chung2016wheeled}
Woojin Chung and Karl Iagnemma.
\newblock Wheeled robots.
\newblock {\em Springer Handbook of Robotics}, pages 575--594, 2016.

\bibitem{choi2016underwater}
Hyun-Taek Choi and Junku Yuh.
\newblock Underwater robots.
\newblock {\em Springer Handbook of Robotics}, pages 595--622, 2016.

\bibitem{santos2020path}
Lu{\'\i}s~C Santos, Filipe~N Santos, EJ~Solteiro Pires, Ant{\'o}nio Valente,
  Pedro Costa, and Sandro Magalh{\~a}es.
\newblock Path planning for ground robots in agriculture: A short review.
\newblock In {\em 2020 IEEE International Conference on Autonomous Robot
  Systems and Competitions (ICARSC)}, pages 61--66. IEEE, 2020.

\bibitem{bawden2017robot}
Owen Bawden, Jason Kulk, Ray Russell, Chris McCool, Andrew English, Feras
  Dayoub, Chris Lehnert, and Tristan Perez.
\newblock Robot for weed species plant-specific management.
\newblock {\em Journal of Field Robotics}, 34(6):1179--1199, 2017.

\bibitem{mcallister2018multi}
Wyatt McAllister, Denis Osipychev, Girish Chowdhary, and Adam Davis.
\newblock Multi-agent planning for coordinated robotic weed killing.
\newblock In {\em 2018 IEEE/RSJ International Conference on Intelligent Robots
  and Systems (IROS)}, pages 7955--7960. IEEE, 2018.

\bibitem{wurman2008coordinating}
Peter~R Wurman, Raffaello D'Andrea, and Mick Mountz.
\newblock Coordinating hundreds of cooperative, autonomous vehicles in
  warehouses.
\newblock {\em AI Magazine}, 29(1):9--9, 2008.

\bibitem{coelho2014thirty}
Leandro~C Coelho, Jean-Fran{\c{c}}ois Cordeau, and Gilbert Laporte.
\newblock Thirty years of inventory routing.
\newblock {\em Transportation Science}, 48(1):1--19, 2014.

\bibitem{queralta2020collaborative}
Jorge~Pena Queralta, Jussi Taipalmaa, Bilge~Can Pullinen, Victor~Kathan Sarker,
  Tuan~Nguyen Gia, Hannu Tenhunen, Moncef Gabbouj, Jenni Raitoharju, and Tomi
  Westerlund.
\newblock Collaborative multi-robot search and rescue: Planning, coordination,
  perception, and active vision.
\newblock {\em IEEE Access}, 8:191617--191643, 2020.

\bibitem{rouvcek2020darpa}
Tom{\'a}{\v{s}} Rou{\v{c}}ek, Martin Pecka, Petr {\v{C}}{\'\i}{\v{z}}ek,
  Tom{\'a}{\v{s}} Pet{\v{r}}{\'\i}{\v{c}}ek, Jan Bayer, Vojt{\v{e}}ch
  {\v{S}}alansk{\`y}, Daniel He{\v{r}}t, Mat{\v{e}}j Petrl{\'\i}k,
  Tom{\'a}{\v{s}} B{\'a}{\v{c}}a, Voj{\v{e}}ch Spurn{\`y}, et~al.
\newblock {DARPA} subterranean challenge: Multi-robotic exploration of
  underground environments.
\newblock In {\em Modelling and Simulation for Autonomous Systems: 6th
  International Conference, MESAS 2019, Palermo, Italy, October 29--31, 2019,
  Revised Selected Papers 6}, pages 274--290. Springer, 2020.

\bibitem{murphy2012projected}
Robin~R Murphy, Joshua Peschel, Clint Arnett, and David Martin.
\newblock Projected needs for robot-assisted chemical, biological,
  radiological, or nuclear ({CBRN}) incidents.
\newblock In {\em 2012 IEEE International Symposium on Safety, Security, and
  Rescue Robotics (SSRR)}, pages 1--4. IEEE, 2012.

\bibitem{hutchinson2019unmanned}
Michael Hutchinson, Cunjia Liu, Paul Thomas, and Wen-Hua Chen.
\newblock Unmanned aerial vehicle-based hazardous materials response:
  Information-theoretic hazardous source search and reconstruction.
\newblock {\em IEEE Robotics \& Automation Magazine}, 27(3):108--119, 2019.

\bibitem{dunbabin2012robots}
Matthew Dunbabin and Lino Marques.
\newblock Robots for environmental monitoring: Significant advancements and
  applications.
\newblock {\em IEEE Robotics \& Automation Magazine}, 19(1):24--39, 2012.

\bibitem{hernandez2012mobile}
Victor Hernandez~Bennetts, Achim~J Lilienthal, Patrick~P Neumann, and Marco
  Trincavelli.
\newblock Mobile robots for localizing gas emission sources on landfill sites:
  is bio-inspiration the way to go?
\newblock {\em Frontiers in Neuroengineering}, 4:20, 2012.

\bibitem{yuan2020maritime}
Haiwen Yuan, Changshi Xiao, Yanfeng Wang, Xin Peng, Yuanqiao Wen, and Qiliang
  Li.
\newblock Maritime vessel emission monitoring by an {UAV} gas sensor system.
\newblock {\em Ocean Engineering}, 218:108206, 2020.

\bibitem{soldan2014towards}
Samuel Soldan, Jochen Welle, Thomas Barz, Andreas Kroll, and Dirk Schulz.
\newblock Towards autonomous robotic systems for remote gas leak detection and
  localization in industrial environments.
\newblock In {\em Field and Service Robotics: Results of the 8th International
  Conference}, pages 233--247. Springer, 2014.

\bibitem{francis2022gas}
Adam Francis, Shuai Li, Christian Griffiths, and Johann Sienz.
\newblock Gas source localization and mapping with mobile robots: A review.
\newblock {\em Journal of Field Robotics}, 39(8):1341--1373, 2022.

\bibitem{merino2012unmanned}
Luis Merino, Fernando Caballero, J~Ramiro Mart{\'\i}nez-de Dios, Iv{\'a}n Maza,
  and An{\'\i}bal Ollero.
\newblock An unmanned aircraft system for automatic forest fire monitoring and
  measurement.
\newblock {\em Journal of Intelligent \& Robotic Systems}, 65:533--548, 2012.

\bibitem{robin2016multi}
Cyril Robin and Simon Lacroix.
\newblock Multi-robot target detection and tracking: taxonomy and survey.
\newblock {\em Autonomous Robots}, 40:729--760, 2016.

\bibitem{parker1995design}
Lynne~E Parker.
\newblock On the design of behavior-based multi-robot teams.
\newblock {\em Advanced Robotics}, 10(6):547--578, 1995.

\bibitem{parker2007distributed}
Lynne~E Parker.
\newblock Distributed intelligence: Overview of the field and its application
  in multi-robot systems.
\newblock In {\em AAAI fall symposium: regarding the intelligence in
  distributed intelligent systems}, pages 1--6, 2007.

\bibitem{schranz2020swarm}
Melanie Schranz, Martina Umlauft, Micha Sende, and Wilfried Elmenreich.
\newblock Swarm robotic behaviors and current applications.
\newblock {\em Frontiers in Robotics and AI}, page~36, 2020.

\bibitem{brambilla2013swarm}
Manuele Brambilla, Eliseo Ferrante, Mauro Birattari, and Marco Dorigo.
\newblock Swarm robotics: a review from the swarm engineering perspective.
\newblock {\em Swarm Intelligence}, 7:1--41, 2013.

\bibitem{golden1987orienteering}
Bruce~L Golden, Larry Levy, and Rakesh Vohra.
\newblock The orienteering problem.
\newblock {\em Naval Research Logistics (NRL)}, 34(3):307--318, 1987.

\bibitem{chao1996team}
I-Ming Chao, Bruce~L Golden, and Edward~A Wasil.
\newblock The team orienteering problem.
\newblock {\em European Journal of Operational Research}, 88(3):464--474, 1996.

\bibitem{gunawan2016orienteering}
Aldy Gunawan, Hoong~Chuin Lau, and Pieter Vansteenwegen.
\newblock Orienteering problem: A survey of recent variants, solution
  approaches and applications.
\newblock {\em European Journal of Operational Research}, 255(2):315--332,
  2016.

\bibitem{vansteenwegen2011orienteering}
Pieter Vansteenwegen, Wouter Souffriau, and Dirk Van~Oudheusden.
\newblock The orienteering problem: A survey.
\newblock {\em European Journal of Operational Research}, 209(1):1--10, 2011.

\bibitem{trevelyan2016robotics}
James Trevelyan, William~R Hamel, and Sung-Chul Kang.
\newblock Robotics in hazardous applications.
\newblock {\em Springer Handbook of Robotics}, pages 1521--1548, 2016.

\bibitem{agmon2017robotic}
Noa Agmon.
\newblock Robotic strategic behavior in adversarial environments.
\newblock In {\em IJCAI}, pages 5106--5110, 2017.

\bibitem{zhou2021multi}
Lifeng Zhou and Pratap Tokekar.
\newblock Multi-robot coordination and planning in uncertain and adversarial
  environments.
\newblock {\em Current Robotics Reports}, 2:147--157, 2021.

\bibitem{prorok2021beyond}
Amanda Prorok, Matthew Malencia, Luca Carlone, Gaurav~S Sukhatme, Brian~M
  Sadler, and Vijay Kumar.
\newblock Beyond robustness: A taxonomy of approaches towards resilient
  multi-robot systems.
\newblock {\em arXiv preprint arXiv:2109.12343}, 2021.

\bibitem{jorgensen2018team}
Stefan Jorgensen, Robert~H Chen, Mark~B Milam, and Marco Pavone.
\newblock The team surviving orienteers problem: routing teams of robots in
  uncertain environments with survival constraints.
\newblock {\em Autonomous Robots}, 42:927--952, 2018.

\bibitem{jorgensen2017matroid}
Stefan Jorgensen, Robert~H Chen, Mark~B Milam, and Marco Pavone.
\newblock The matroid team surviving orienteers problem: Constrained routing of
  heterogeneous teams with risky traversal.
\newblock In {\em 2017 IEEE/RSJ International Conference on Intelligent Robots
  and Systems (IROS)}, pages 5622--5629. IEEE, 2017.

\bibitem{jorgensen2024matroid}
Stefan Jorgensen and Marco Pavone.
\newblock The matroid team surviving orienteers problem and its variants:
  Constrained routing of heterogeneous teams with risky traversal.
\newblock {\em The International Journal of Robotics Research}, 43(1):34--52,
  2024.

\bibitem{di2022foraging}
Kai Di, Yifeng Zhou, Fuhan Yan, Jiuchuan Jiang, Shaofu Yang, and Yichuan Jiang.
\newblock A foraging strategy with risk response for individual robots in
  adversarial environments.
\newblock {\em ACM Transactions on Intelligent Systems and Technology (TIST)},
  13(5):1--29, 2022.

\bibitem{shi2023robust}
Guangyao Shi, Lifeng Zhou, and Pratap Tokekar.
\newblock Robust multiple-path orienteering problem: Securing against
  adversarial attacks.
\newblock {\em IEEE Transactions on Robotics}, 2023.

\bibitem{korngut2023multi}
Yair Korngut and Noa Agmon.
\newblock Multi-robot heterogeneous adversarial coverage.
\newblock In {\em 2023 International Symposium on Multi-Robot and Multi-Agent
  Systems (MRS)}, pages 100--106. IEEE, 2023.

\bibitem{yehoshua2016robotic}
Roi Yehoshua, Noa Agmon, and Gal~A Kaminka.
\newblock Robotic adversarial coverage of known environments.
\newblock {\em The International Journal of Robotics Research},
  35(12):1419--1444, 2016.

\bibitem{liu2021distributed}
Jun Liu, Lifeng Zhou, Pratap Tokekar, and Ryan~K Williams.
\newblock Distributed resilient submodular action selection in adversarial
  environments.
\newblock {\em IEEE Robotics and Automation Letters}, 6(3):5832--5839, 2021.

\bibitem{zhou2022distributed}
Lifeng Zhou, Vasileios Tzoumas, George~J Pappas, and Pratap Tokekar.
\newblock Distributed attack-robust submodular maximization for multirobot
  planning.
\newblock {\em IEEE Transactions on Robotics}, 38(5):3097--3112, 2022.

\bibitem{mayya2022adaptive}
Siddharth Mayya, Ragesh~K Ramachandran, Lifeng Zhou, Vinay Senthil, Dinesh
  Thakur, Gaurav~S Sukhatme, and Vijay Kumar.
\newblock Adaptive and risk-aware target tracking for robot teams with
  heterogeneous sensors.
\newblock {\em IEEE Robotics and Automation Letters}, 7(2):5615--5622, 2022.

\bibitem{zhou2018resilient}
Lifeng Zhou, Vasileios Tzoumas, George~J Pappas, and Pratap Tokekar.
\newblock Resilient active target tracking with multiple robots.
\newblock {\em IEEE Robotics and Automation Letters}, 4(1):129--136, 2018.

\bibitem{schwager2017multi}
Mac Schwager, Philip Dames, Daniela Rus, and Vijay Kumar.
\newblock A multi-robot control policy for information gathering in the
  presence of unknown hazards.
\newblock In {\em Robotics Research: The 15th International Symposium ISRR},
  pages 455--472. Springer, 2017.

\bibitem{shapira2015path}
Yaniv Shapira and Noa Agmon.
\newblock Path planning for optimizing survivability of multi-robot formation
  in adversarial environments.
\newblock In {\em 2015 IEEE/RSJ International Conference on Intelligent Robots
  and Systems (IROS)}, pages 4544--4549. IEEE, 2015.

\bibitem{huang2019survey}
Li~Huang, MengChu Zhou, Kuangrong Hao, and Edwin Hou.
\newblock A survey of multi-robot regular and adversarial patrolling.
\newblock {\em IEEE/CAA Journal of Automatica Sinica}, 6(4):894--903, 2019.

\bibitem{pardalos2017non}
Panos~M Pardalos, Antanas {\v{Z}}ilinskas, and Julius {\v{Z}}ilinskas.
\newblock {\em Non-convex multi-objective optimization}.
\newblock Springer, 2017.

\bibitem{branke2008multiobjective}
J{\"u}rgen Branke, Kalyanmoy~Deb Deb, Kaisa Miettinen, and Roman
  S\l{}owi\'{n}ski.
\newblock {\em Multiobjective optimization: Interactive and evolutionary
  approaches}.
\newblock Springer Science \& Business Media, 2008.

\bibitem{iredi2001bi}
Steffen Iredi, Daniel Merkle, and Martin Middendorf.
\newblock Bi-criterion optimization with multi colony ant algorithms.
\newblock In {\em Evolutionary Multi-Criterion Optimization: First
  International Conference, EMO 2001 Zurich, Switzerland, March 7--9, 2001
  Proceedings 1}, pages 359--372. Springer, 2001.

\bibitem{clark1991first}
John Clark and Derek~Allan Holton.
\newblock {\em A First Look at Graph Theory}.
\newblock World Scientific, 1991.

\bibitem{graphtheory2}
Santosh~Kumar Yadav.
\newblock {\em Advanced Graph Theory}.
\newblock Springer, 2023.

\bibitem{wilson1979introduction}
Robin~J Wilson.
\newblock {\em Introduction to graph theory}.
\newblock Pearson Education India, 1979.

\bibitem{tang2023poisson}
Wenpin Tang and Fengmin Tang.
\newblock The {P}oisson binomial distribution—old \& new.
\newblock {\em Statistical Science}, 38(1):108--119, 2023.

\bibitem{dorigo2006ant}
Marco Dorigo, Mauro Birattari, and Thomas Stutzle.
\newblock Ant colony optimization.
\newblock {\em IEEE Computational Intelligence Magazine}, 1(4):28--39, 2006.

\bibitem{bonabeau1999swarm}
Eric Bonabeau, Marco Dorigo, and Guy Theraulaz.
\newblock {\em {Swarm Intelligence: From Natural to Artificial Systems}}.
\newblock Number~1. Oxford university press, 1999.

\bibitem{blum2005ant}
Christian Blum.
\newblock Ant colony optimization: Introduction and recent trends.
\newblock {\em Physics of Life Reviews}, 2(4):353--373, 2005.

\bibitem{bonabeau2000inspiration}
Eric Bonabeau, Marco Dorigo, and Guy Theraulaz.
\newblock Inspiration for optimization from social insect behaviour.
\newblock {\em Nature}, 406(6791):39--42, 2000.

\bibitem{ke2008ants}
Liangjun Ke, Claudia Archetti, and Zuren Feng.
\newblock Ants can solve the team orienteering problem.
\newblock {\em Computers \& Industrial Engineering}, 54(3):648--665, 2008.

\bibitem{chen2015multiobjective}
Yu-Han Chen, Wei-Ju Sun, and Tsung-Che Chiang.
\newblock Multiobjective orienteering problem with time windows: An ant colony
  optimization algorithm.
\newblock In {\em 2015 Conference on Technologies and Applications of
  Artificial Intelligence (TAAI)}, pages 128--135. IEEE, 2015.

\bibitem{verbeeck2017time}
C{\'e}dric Verbeeck, Pieter Vansteenwegen, and El-Houssaine Aghezzaf.
\newblock The time-dependent orienteering problem with time windows: a fast ant
  colony system.
\newblock {\em Annals of Operations Research}, 254:481--505, 2017.

\bibitem{sohrabi2021acs}
Somayeh Sohrabi, Koorush Ziarati, and Morteza Keshtkaran.
\newblock {ACS-OPHS}: Ant colony system for the orienteering problem with hotel
  selection.
\newblock {\em EURO Journal on Transportation and Logistics}, 10:100036, 2021.

\bibitem{chen2022environment}
Jianhui Chen, Shiqi Zeng, and Xiaoli Xu.
\newblock Environment-aware path planning for {UAV}-assisted search and rescue.
\newblock In {\em 2022 14th International Conference on Wireless Communications
  and Signal Processing (WCSP)}, pages 922--927. IEEE, 2022.

\bibitem{montemanni2011enhanced}
Roberto Montemanni, D~Weyland, and LM~Gambardella.
\newblock An enhanced ant colony system for the team orienteering problem with
  time windows.
\newblock In {\em 2011 international symposium on computer science and
  society}, pages 381--384. IEEE, 2011.

\bibitem{gavalas2014survey}
Damianos Gavalas, Charalampos Konstantopoulos, Konstantinos Mastakas, and
  Grammati Pantziou.
\newblock A survey on algorithmic approaches for solving tourist trip design
  problems.
\newblock {\em Journal of Heuristics}, 20(3):291--328, 2014.

\bibitem{dang2013effective}
Duc-Cuong Dang, Rym~Nesrine Guibadj, and Aziz Moukrim.
\newblock An effective pso-inspired algorithm for the team orienteering
  problem.
\newblock {\em European Journal of Operational Research}, 229(2):332--344,
  2013.

\bibitem{chao1996fast}
I-Ming Chao, Bruce~L Golden, and Edward~A Wasil.
\newblock A fast and effective heuristic for the orienteering problem.
\newblock {\em European Journal of Operational Research}, 88(3):475--489, 1996.

\bibitem{butt1994heuristic}
Steven~E Butt and Tom~M Cavalier.
\newblock A heuristic for the multiple tour maximum collection problem.
\newblock {\em Computers \& Operations Research}, 21(1):101--111, 1994.

\bibitem{evison2008combined}
Sophie~EF Evison, Owen~L Petchey, Andrew~P Beckerman, and Francis~LW Ratnieks.
\newblock Combined use of pheromone trails and visual landmarks by the common
  garden ant {L}asius niger.
\newblock {\em Behavioral Ecology and Sociobiology}, 63:261--267, 2008.

\bibitem{czaczkes2015trail}
Tomer~J Czaczkes, Christoph Gr{\"u}ter, and Francis~LW Ratnieks.
\newblock Trail pheromones: an integrative view of their role in social insect
  colony organization.
\newblock {\em Annual Review of Entomology}, 60:581--599, 2015.

\bibitem{robinson2005no}
Elva~JH Robinson, Duncan~E Jackson, Mike Holcombe, and Francis~LW Ratnieks.
\newblock ‘no entry’ signal in ant foraging.
\newblock {\em Nature}, 438(7067):442--442, 2005.

\bibitem{deneubourg1983probabilistic}
Jean-Louis Deneubourg, Jacques~M Pasteels, and Jean-Claude Verhaeghe.
\newblock Probabilistic behaviour in ants: a strategy of errors?
\newblock {\em Journal of Theoretical Biology}, 105(2):259--271, 1983.

\bibitem{goss1989self}
Simon Goss, Serge Aron, Jean-Louis Deneubourg, and Jacques~Marie Pasteels.
\newblock Self-organized shortcuts in the {A}rgentine ant.
\newblock {\em Naturwissenschaften}, 76(12):579--581, 1989.

\bibitem{beckers1993modulation}
Ralph Beckers, Jean-Louis Deneubourg, and Simon Goss.
\newblock Modulation of trail laying in the ant {L}asius niger (hymenoptera:
  Formicidae) and its role in the collective selection of a food source.
\newblock {\em Journal of Insect Behavior}, 6:751--759, 1993.

\bibitem{david2009trail}
E~David~Morgan.
\newblock Trail pheromones of ants.
\newblock {\em Physiological Entomology}, 34(1):1--17, 2009.

\bibitem{knaden2016sensory}
Markus Knaden and Paul Graham.
\newblock The sensory ecology of ant navigation: from natural environments to
  neural mechanisms.
\newblock {\em Annual Review of Entomology}, 61:63--76, 2016.

\bibitem{jackson2006communication}
Duncan~E Jackson and Francis~LW Ratnieks.
\newblock Communication in ants.
\newblock {\em Current Biology}, 16(15):R570--R574, 2006.

\bibitem{van2011temperature}
Louise Van~Oudenhove, Elise Billoir, Rapha{\"e}l Boulay, Carlos Bernstein, and
  Xim Cerd{\'a}.
\newblock Temperature limits trail following behaviour through pheromone decay
  in ants.
\newblock {\em Naturwissenschaften}, 98:1009--1017, 2011.

\bibitem{czaczkes2013ant}
Tomer~J Czaczkes, Christoph Gr{\"u}ter, Laura Ellis, Elizabeth Wood, and
  Francis~LW Ratnieks.
\newblock Ant foraging on complex trails: route learning and the role of trail
  pheromones in lasius niger.
\newblock {\em Journal of Experimental Biology}, 216(2):188--197, 2013.

\bibitem{wendt2020negative}
Stephanie Wendt, Nico Kleinhoelting, and Tomer~J Czaczkes.
\newblock Negative feedback: ants choose unoccupied over occupied food sources
  and lay more pheromone to them.
\newblock {\em Journal of The Royal Society Interface}, 17(163):20190661, 2020.

\bibitem{deneubourg1990self}
J~L Deneubourg, Serge Aron, Simon Goss, and Jacques~M Pasteels.
\newblock The self-organizing exploratory pattern of the {A}rgentine ant.
\newblock {\em Journal of Insect Behavior}, 3:159--168, 1990.

\bibitem{shiraishi2019diverse}
Masashi Shiraishi, Rito Takeuchi, Hiroyuki Nakagawa, Shin~I Nishimura, Akinori
  Awazu, and Hiraku Nishimori.
\newblock Diverse stochasticity leads a colony of ants to optimal foraging.
\newblock {\em Journal of Theoretical Biology}, 465:7--16, 2019.

\bibitem{deneubourg1986random}
Jean~L Deneubourg, Serge Aron, SAPJM Goss, JM~Pasteels, and Guido Duerinck.
\newblock Random behaviour, amplification processes and number of participants:
  how they contribute to the foraging properties of ants.
\newblock {\em Physica D: Nonlinear Phenomena}, 22(1-3):176--186, 1986.

\bibitem{dussutour2009noise}
Audrey Dussutour, Madeleine Beekman, Stamatios~C Nicolis, and Bernd Meyer.
\newblock Noise improves collective decision-making by ants in dynamic
  environments.
\newblock {\em Proceedings of the Royal Society B: Biological Sciences},
  276(1677):4353--4361, 2009.

\bibitem{edelstein1995trail}
Leah Edelstein-Keshet, James Watmough, and G~Bard Ermentrout.
\newblock Trail following in ants: individual properties determine population
  behaviour.
\newblock {\em Behavioral Ecology and Sociobiology}, 36:119--133, 1995.

\bibitem{bonabeau1997self}
Eric Bonabeau, Guy Theraulaz, Jean-Louls Deneubourg, Serge Aron, and Scott
  Camazine.
\newblock Self-organization in social insects.
\newblock {\em Trends in Ecology \& Evolution}, 12(5):188--193, 1997.

\bibitem{watmough1995modelling}
James Watmough and Leah Edelstein-Keshet.
\newblock Modelling the formation of trail networks by foraging ants.
\newblock {\em Journal of Theoretical Biology}, 176(3):357--371, 1995.

\bibitem{dorigo1996ant}
Marco Dorigo, Vittorio Maniezzo, and Alberto Colorni.
\newblock Ant system: optimization by a colony of cooperating agents.
\newblock {\em IEEE Transactions on Systems, Man, and Cybernetics, Part B
  (Cybernetics)}, 26(1):29--41, 1996.

\bibitem{cao2015using}
Yongtao Cao, Byran~J Smucker, and Timothy~J Robinson.
\newblock On using the hypervolume indicator to compare {P}areto fronts:
  Applications to multi-criteria optimal experimental design.
\newblock {\em Journal of Statistical Planning and Inference}, 160:60--74,
  2015.

\bibitem{guerreiro2020hypervolume}
Andreia~P Guerreiro, Carlos~M Fonseca, and Lu{\'\i}s Paquete.
\newblock The hypervolume indicator: Problems and algorithms.
\newblock {\em arXiv preprint arXiv:2005.00515}, 2020.

\bibitem{yu2014correlated}
Jingjin Yu, Mac Schwager, and Daniela Rus.
\newblock Correlated orienteering problem and its application to informative
  path planning for persistent monitoring tasks.
\newblock In {\em 2014 IEEE/RSJ International Conference on Intelligent Robots
  and Systems}, pages 342--349. IEEE, 2014.

\bibitem{asghar2023risk}
Ahmad~Bilal Asghar, Guangyao Shi, Nare Karapetyan, James Humann, Jean-Paul
  Reddinger, James Dotterweich, and Pratap Tokekar.
\newblock Risk-aware recharging rendezvous for a collaborative team of {UAV}s
  and {UGV}s.
\newblock In {\em 2023 IEEE International Conference on Robotics and Automation
  (ICRA)}, pages 5544--5550. IEEE, 2023.

\bibitem{khuller2011fill}
Samir Khuller, Azarakhsh Malekian, and Juli{\'a}n Mestre.
\newblock To fill or not to fill: The gas station problem.
\newblock {\em ACM Transactions on Algorithms (TALG)}, 7(3):1--16, 2011.

\bibitem{liao2016electric}
Chung-Shou Liao, Shang-Hung Lu, and Zuo-Jun~Max Shen.
\newblock The electric vehicle touring problem.
\newblock {\em Transportation Research Part B: Methodological}, 86:163--180,
  2016.

\bibitem{yu2019coverage}
Kevin Yu, Jason~M O’Kane, and Pratap Tokekar.
\newblock Coverage of an environment using energy-constrained unmanned aerial
  vehicles.
\newblock In {\em 2019 International Conference on Robotics and Automation
  (ICRA)}, pages 3259--3265. IEEE, 2019.

\bibitem{majumdar2020should}
Anirudha Majumdar and Marco Pavone.
\newblock How should a robot assess risk? towards an axiomatic theory of risk
  in robotics.
\newblock In {\em Robotics Research: The 18th International Symposium ISRR},
  pages 75--84. Springer, 2020.

\bibitem{burton1992instance}
Didier Burton and Ph~L Toint.
\newblock On an instance of the inverse shortest paths problem.
\newblock {\em Mathematical Programming}, 53:45--61, 1992.

\end{thebibliography}
\bibliographystyle{unsrt}

\end{document}